\documentclass[10pt,a4paper, table]{article}
\usepackage[latin1]{inputenc}
\usepackage{amsmath}
\usepackage{amsfonts}
\usepackage{amssymb}
\usepackage{makeidx}
\usepackage{graphicx}
\usepackage{todonotes}
\usepackage{lipsum}
\usepackage{tabularx}
\usepackage{float}
\usepackage{multirow}
\usepackage{xcolor}
\usepackage{upgreek}
\definecolor{grey}{RGB}{211,211,211}

\providecommand{\keywords}[1]
{
	\small	
	\textbf{Keywords---} #1
}

\author{
	Rik van Leeuwen$^{1,2}$ $\bullet$ Ger Koole$^{2}$ \\
	\texttt{\scriptsize\sffamily $^{1}$Ireckonu, Olympisch Stadion 43, 1076DE, Amsterdam, The Netherlands} \\
	\texttt{\scriptsize\sffamily $^{2}$Department of Mathematics, Vrije Universiteit, De Boelelaan 1111, 1081HV Amsterdam, The Netherlands} \\
	\texttt{\scriptsize\sffamily $^{1}$rik@ireckonu.com $\bullet$ $^{2}$ger.koole@vu.nl} \\
}
\title{Data-Driven Market Segmentation in Hospitality Using Unsupervised Machine Learning}
\date{\today}

\begin{document}
	
	\maketitle
	
	\begin{abstract}
		Within hospitality, marketing departments use segmentation to create tailored strategies to ensure personalized marketing. This study provides a data-driven approach by segmenting guest profiles via hierarchical clustering, based on an extensive set of features. The industry requires understandable outcomes that contribute to adaptability for marketing departments to make data-driven decisions and ultimately driving profit. 
		
		A marketing department specified a business question that guides the unsupervised machine learning algorithm. Features of guests change over time; therefore, there is a probability that guests transition from one segment to another. The purpose of the study is to provide steps in the process from raw data to actionable insights, which serve as a guideline for how hospitality companies can adopt an algorithmic approach. 
	\end{abstract}
	
	
	\section{Introduction}
	\label{sec: introduction}
	
	Each business selling a product to a larger audience or customer base has the opportunity to create segments or clusters. Segmentation divides and defines a customer base into groups based on similar behaviour (Tynan \& Drayton, 1987). The main goal of segmentation is to form a strategy tailored to a group, which supports corporate objective and growth. The customer base within hospitality consists of the guests who are interested in staying or who have stayed at the hotel.
	
	Segments are defined by similarities which are based on attributes or features of a geographical, behavioral and/or demographical type (Jurowski \& Reich, 2000). The way to identify segments depends on several factors, such as the available data, employee skills, and resources. In practice, there are two main ways to identify segments: rule-based (qualitative) or algorithmic-based (quantitative) (Tu, Dong et al., 2010). Rule-based approaches are formulated by defining guest personas based on a handful of attributes and gut feeling (Sinha, 2003). These guest personas allow the hotel to develop tailored marketing strategies. Pitfalls for rule-based approaches are that not all guests belong to a persona, personas are outdated, or they are based on a small set of attributes (Tu, Dong et al., 2010). Having large amounts of data can be a valid reason to apply algorithms to find new segments evaluated on an extended set of features.
	
	Data-driven segmentation via algorithms, known as unsupervised machine learning, guarantees an outcome regardless of the quality of the data or the features included. To evaluate the outcome of the algorithm, a business question or research problem needs to be defined, which means there is an end in mind (Arunachalam \& Kumar, 2018). The features of a guest, which serve as the input of an algorithm, need to reflect and allow for an answer to marketing questions such as: which guests to market and who is most valuable to the brand?
	
	The selection of features is an iterative and creative process until the outcome allows for an to answer to the underlying business question. Data exploration and pre-processing are steps within this iterative process (Arunachalam \& Kumar, 2018). These steps overcome potential data issues, ensure quality, and allow for the selection of features. Previous research shows that bad data or studies without a defined goal lead to bad analytical results, a process often referred to as the ``garbage in, garbage'' out principle (Baesens, Mues, Martens, \& Vanthienen, 2009). If the underlying business question is not properly defined, these steps can help finalize the main questions. A critical component within these steps is the involvement of the end user, which is the marketing department. The interpretation of the algorithmic segmentation can be challenging (Mariani, 2018), which directly impacts the willingness of adoption by marketing departments.
	
	In practice, the hospitality industry applies rule-based systems to segment guests instead of algorithmic approaches. In this study, profiles are segmented by dividing guests based on similar characteristics as a result of an algorithmic approach. The underlying business question is formulated together with the marketing department of a single hotel property based in Amsterdam, the Netherlands. The main purpose of segmenting guests is to obtain a balanced separation of guests based on the lifetime value and the channel used to make a reservation. This split of 170,000 guests, as a result of the unsupervised machine learning method, guides the marketing department in formulating a tailored marketing strategy. Analysis over time showcases how and why profiles transition over time.
	
	This study shows, via a case study, how a hospitality company can make use of an unsupervised machine learning method to segment guests, which influences their marketing strategy for specific groups that leads to a competitive advantage. Unique insights are presented about segments over time so marketeers have direct starting points to influence profiles transitioning from one segment to another. This paper  is a guideline for hospitality to transition to data-driven market segmentation via an algorithmic approach due to a representable dataset. The guideline is created based on a literature review in Section 2, a description of the data in Section 3, and an extensive data analysis in Section 4. The steps of pre-processing the data and feature engineering are described in Section 5. The model and configuration are presented  in Section 6, followed by results of the case study in Section 7. The conclusion and discussion are presented in Section 8.

	\section{Literature Review}
	\label{sec: literaturereview}
	
	Marketeers create strategies or policies serving the right approach for the right guest, typically to achieve conversion or retention. Marketing departments are restricted by time and financial resources. Generally speaking, there exist two extremes of strategies: a single strategy for all guests versus individual, focussed strategies. In the middle of these two extremes, there is a strategy for groups. With available resources, marketeers strive for as personalized a marketing strategy as possible because of the effectiveness of such a strategy (Bleier \& Eisenbeiss, 2015).
	
	\newpage
	
	Personalized marketing transforms a transactional-oriented organization to a relationship- oriented organization (Jain \& Jain, 2005). The cost of gaining a new guest can be five times higher than retaining one (McIlroy \& Barnett, 2000). Effectively identifying guest behaviour and interests increases satisfaction and consequently builds a relationship. Marketing segmentation, as part of a marketing strategy, leans towards a personalized approach.
	
	Market segmentation is a creative approach which categorises guests into smaller groups based on shared characteristics (Tynan \& Drayton, 1987). Understanding characteristics and behaviour of groups creates the opportunity of executing a more personalized marketing strategy. The result of this creative approach should meet six criteria ? identifiability, substantiality, accessibility, stability, responsiveness, and actionability ? which determine the effectiveness, the stability, and profitability of marketing strategies (Wedel \& Kamakura, 1999).
	
	The effectiveness and efficiency of a marketing budget are increased by applying market segmentation (Jurowski \& Reich, 2000). Segmenting guests is generally achieved via (1) a priori segmentation; or (2) post-hoc segmentation. The main difference between these approaches is (1) using known criteria and (2) using characteristics to describe guests. Post-hoc, known as data-driven, segmentation leads to a competitive advantage since the most common approach is a priori (Dolnicar, 2002).
	
	Segments are required to be identifiable, which is dependent on the available data, the sample size, and the quality of available attributes or features per guest (Dolnicar, 2002). The attributes, which distinguishes guests, are derived from the research objective or problem. For example, if the main purpose is to identify profitable guests only transactional data is required during a stay. The next stage is to add data points, e.g. booking channel, gender, age, to understand where to spend the marketing budget.
	
	There is a wide range of algorithms available to segment unlabelled guests, known as cluster analysis (Jurowski \& Reich, 2000). The advantage of applying algorithms is that thousands of guests are evaluated objectively by considering all features. An algorithm takes guest-level data as input to divide the data set into subgroups. Afterwards, a manual review is required to provide context about these subgroups based on the distinction of characteristics (Dolnicar, 2002).
	
	There are two types of algorithmic clustering methods: (1) hierarchical and (2) non-hierarchical. The main difference between these two types is that non-hierarchical clustering requires the number of clusters as an input variable, while hierarchical does not require this (Jurowski \& Reich, 2000). Because of the research objective, a hierarchical approach is the focus in this study as the creative process divides guests on a high level where the number of clusters is not known upfront.
	
	When dealing with unlabelled data, a new challenge arises that is as old as the clustering method itself: defining the number of clusters (Thorndike, 1953). The number of clusters needs to be defined before running the unsupervised machine learning algorithm, which strongly influences the solution. Various approaches, i.e. using heuristics, subjective opinions, or a combination of those two, have been proposed to tackle this problem. However, no standard procedure has been formulated for obtaining the optimal number of clusters. A recommended approach is to rerun the algorithm with different input parameters influenced by either algorithmic criteria, e.g. stability, or by corporate criteria given by management (Dolnicar, 2002).
	
	A common way for marketeers to reach these market segments defined by a set of features is via direct email marketing (Marinova, 2002). In hospitality, email information, such as address and opt-in marketing consent, is part of the profile data. Characteristics of guests, such as preferences, purchasing history, and intentions, need to be understood and implemented into a direct email message to shift towards a relationship-oriented organization. Marketeers, having obtained knowledge about guest characteristics, create targeted lists of customers for the delivery of personalized marketing (Stone \& Mason, 1997).
	
	\section{Data}
	\label{sec: data}
	
	The data for this study originates from a Property Management System (PMS), a system used in the hospitality industry to register and regulate reservations. Reservations, profiles, and folios from 2015 until 2019 of a single-property hotel based in Amsterdam, the Netherlands are included. These data points are structured such that profiles, reservations, and folios (record of stay finances) are separate tables. Figure~7 in Appendix A provides a complete overview of the data, which is a representative structure for the hospitality industry. This dataset does not contain any missing data fields. Personally identifiable information is excluded due to the General Data Protection Regulation (GDPR)\footnote{1Regulation (EU) 2016/679 of the European Parliament and of the Council of 27 April 2016 on the protection of natural persons with regard to the processing of personal data and on the free movement of such data, and repealing Directive 95/46/EC (General Data Protection Regulation) (Text with EEA relevance), [2016] OJ L 119/1.}.
	
	In this study, a profile is considered the centre of the universe. Therefore, each data point must be associated with the same person, a profile within the available data set. However, when a person has two profiles within the dataset, these data points are not associated with the same person. A person can have two profiles by making a reservation via different companies for example. A golden profile is generated by applying match and merge on a profile level. Match and merge is a rule-based system that compares personal data points of profiles ? such as first name, name, email home address, address, phone number ? based on an exact match as well as a phonetic key. A golden profile combines all necessary guest information with fundamental insights (e.g. sentiment of online reviews, engagement with hotel facilities) and metrics (e.g. the total amount spent, recurrence) unique to a profile. Such a profile truly is the centre of the universe in a data set such as this.
	
	Additional attributes are added based on the data from the reservations table. One of these points is lead time, which is the number of days between the booking date and the arrival date. Another attribute is the length of stay, which is the number of days between the arrival date and the departure date.
	
	Since the data originates from a PMS, attributes of reservations or folios may contain a range of distinct values dependent on the configuration. By adding a classification of attributes, additional context is presented to an algorithm. Examples  of attributes that can be classified are sources, transactional codes, and rate plans. The source of a reservation can be referred to as \textit{direct} or \textit{indirect}. \textit{Direct} classified reservations have been produced via the website of the hotel, phone, or walk-in. \textit{Indirect} reservations originate from the Global Distribution System (GDS) or an Online Travel Agency (OTA) for example. Revenues of transactional codes are mapped to either \textit{Room}, \textit{Ancillary}, or \textit{Other}. Ancillary revenue is revenue derived from services or goods, such as a restaurant or spa. Transactional codes \textit{Other} are taxes and tips, which are not considered as revenue and therefore excluded from the folios. This classification of attributes contributes to the comparison and understandability of segments.
	
	\section{Data Analysis}
	\label{sec: data analysis}
	
	A data analysis is executed to gain knowledge about attributes that distinguish be- tween guest behaviours and reasons to travel. The data set contains around $170,000$ profiles, $230,000$ reservations and $870,000$ folios created between $2015$ and $2019$. Reservations are given a distinct status: stay, cancelled, or no-show. For reservations overall $82.1\%$ are marked as stay, $16.3\%$ as cancelled, and $1.6\%$ as no-show. Focusing on individual room nights, $80.7\%$ of the nights are marked as stay, $17.8\%$ as cancelled, and $1.5\%$ as no-show. In terms of the transactional code classification, $77.0\%$ is generated by \textit{Room revenue} and the other $23.0\%$ is generated by \textit{Ancillary revenue}.
	
	The number of profiles grows over time with new guests making reservations. Each year around $34,000$ new profiles are registered, and $82.2\%$ of those new profiles had a `stay' reservation. The other $17.8\%$ of these profiles reserved and cancelled or were a no-show. Based on all profiles, $14.3\%$ are marked as repeat guests, i.e., the number of `stay' reservations is larger than one. On average, repeat guests spend $16.7\%$ more than guests who stay once. These repeat profiles produced $36.2\%$ of the reservations and $35.1\%$ of the total revenue.
	
	Guest are able to retrieve a loyalty status from a hotel, indicated with the feature \textit{LoyaltyBinary}. This loyalty status is granted to $1.17\%$ of the profiles based on their life-time value or an important/celebrity status. This type of feature is generally speaking only available on a chain level instead of an individual property level. The total amount spent is higher ($4.6$ times) than that of a profile without a loyalty status, which is primarily caused by a higher number of stays ($4.9$ times).
	
	Figure 1 shows two repeat percentages: one taking into account all reservations (black line) and another taking into account the last 365 days (grey line). Only reservations and the associated profiles with the status Historic are selected. The first 365 days, seen as the warm-up period, are not displayed. There is a recurring yearly pattern with a peak in the repeat percentage during November until February. Both lines show a similar pattern with an average difference of $8.21$ percentage point.
	
	\begin{figure}[H]
		\centering
		\includegraphics[width=1\linewidth]{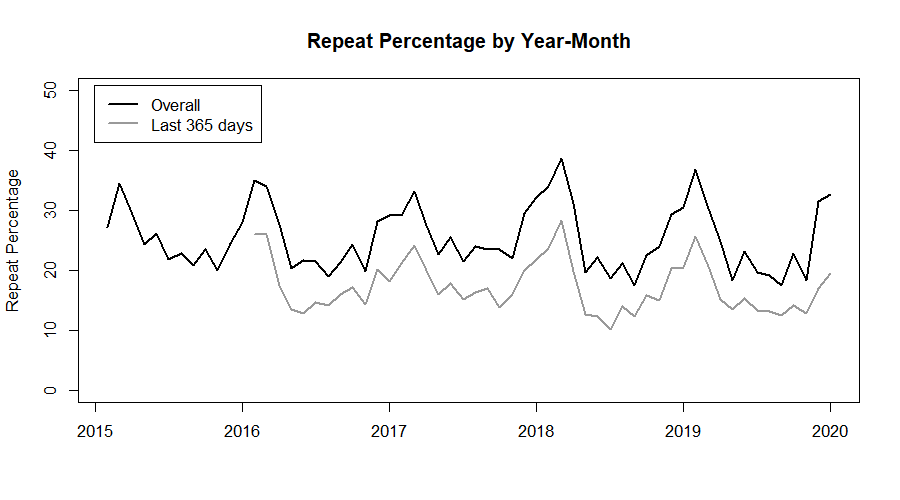}
		\caption{Repeat percentage per arrival date of reservation with status historic aggregated by year-month}
		\label{fig:EDA - Repeat}
	\end{figure}

	Table 1 shows the retention rate of profiles with an association to a group, company, and/or agency. This table only includes profiles ($137,800$) with first-time stays between $2015$ and $2019$. The overall retention rate of these profiles is $8.55\%$. Profiles that are not related to a group or to a company have the highest percentage of becoming a repeat profile. Profiles that are related to a group and to an agency have the lowest retention rate, regardless of whether a company is associated to a profile.
	
	\begin{table}[]
		\begin{tabular}{ccc|cc}
			\textbf{Group} & \textbf{Company} & \textbf{Agency} & \textbf{No. of Profiles} & \textbf{Retention Percentage} \\ \hline
			Yes            & No               & Yes             & 16,798                   & 2.40 \%                       \\
			Yes            & No               & No              & 13,507                   & 5.44 \%                       \\
			Yes            & Yes              & Yes             & 4,665                    & 3.90 \%                       \\
			Yes            & Yes              & No              & 20,597                   & 9.94 \%                       \\
			No             & No               & No              & 41,301                   & 10.80 \%                      \\
			No             & No               & Yes             & 32,132                   & 7.62 \%                       \\
			No             & Yes              & No              & 6,190                    & 17.54\%                       \\
			No             & Yes              & Yes             & 2,610                    & 16.32\%                       \\ \hline
			-              & -                & -               & 137,800                  & 8.55 \%                       \\ \hline
		\end{tabular}
		\caption{Profile distribution of becoming repeat based on group, company, and agency attributes of reservations}
	\end{table}

	Hotels prefer guests to book via \textit{direct} channels because of the commission that needs be to paid to \textit{indirect} channels. The data set contains $11$ distinct channels. From a reservations perspective, $44.49\%$ of the reservations came from direct and $55.51\%$ came from indirect channels. From a profile perspective, $40.90\%$ of the profiles booked via direct channels only, $55.44\%$ via indirect channels only, and $3.66\%$ used both types of channels to book a room. The average amount spent is $12.84\%$ higher from guests who book via \textit{direct} channels compared to guests who book via \textit{indirect} channels, without subtracting the commission. Only taking into account guests who come for the first time, the average amount spent is $7.06\%$ higher, again not taking into account the commission.
	
	Guests travel for different reasons, two of the most common being leisure and business. One of the characteristics is when a guest arrives and departs based on  the day of the week. A business guest tends to arrive and depart during week days, Monday to Thursday, while a leisure guest tends to arrive on weekend days, Friday to Sunday. This does not mean that if a reservation ticks one of these boxes, it is classified as a leisure or business guest. Table 2 shows the distribution of reservations with the status \textit{historic} per year matching the \textit{week} criterion, \textit{weekend} criterion, and \textit{other} if it was neither of the previous options. Week stay reservations fluctuate over the years, while the weekend stay reservation remain consistent over the years.
	
	\begin{table}[H]
		\centering
		\begin{tabular}{l|ccccc}
			\textbf{Stay}    & \textbf{2015} & \textbf{2016} & \textbf{2017} & \textbf{2018} & \textbf{2019} \\ \hline
			\textbf{Week}    & 45.8          & 49.1          & 48.2          & 43.9          & 49.3          \\
			\textbf{Weekend} & 31.0          & 32.0          & 31.5          & 33.1          & 32.4          \\
			\textbf{Other}   & 23.2          & 18.9          & 20.3          & 23.0          & 18.3         
		\end{tabular}
		\caption{Reservations divided by arrival}
	\end{table}
	
	\textit{Lead time} indicates whether a reservation is booked far in advance (early-bird) or just before arrival (last-minute). A guest is able to book a room 365 days in advance. Figure 2 shows the lead time distribution of all reservations. The average lead time is 31 days, indicated by the grey dotted line. $19.8\%$ of all reservations are made with a lead time larger than 45 days, and around $15.2\%$ of reservations are made within 3 days before arrival. Additional attributes are constructed based on the lead time feature, which is dictated by business logic, see Section 5.1.
	
	Taking into account all reservations, guests make a reservation for an average of $2.15$ nights. Of the guest who book a single night, which is $40.0\%$ of the reservations, $57.2\%$ stay during the week and $42.8\%$ during the weekend. $13.2\%$ of all reservations stay longer than 3 nights.
	
	\begin{figure}[]
		\centering
		\includegraphics[width=1\linewidth]{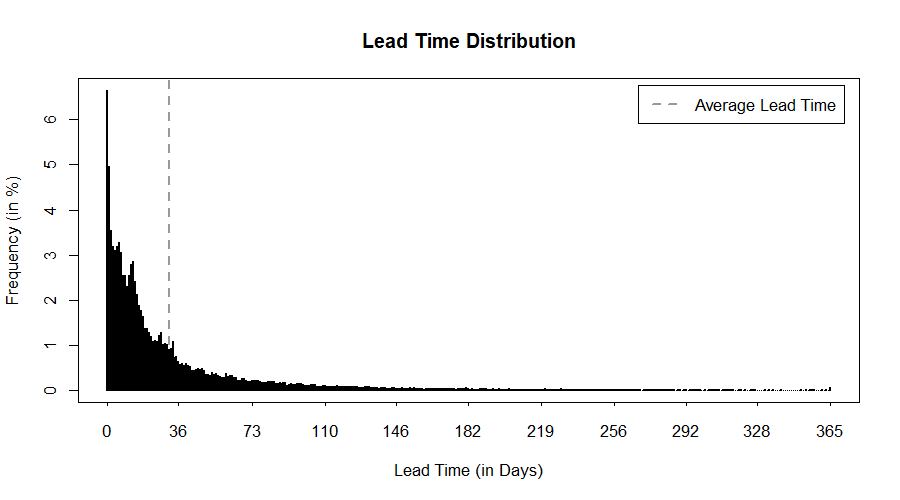}
		\caption{Lead time distribution of all reservations}
		\label{fig:EDA - Lead Time}
	\end{figure}
	
	\section{Data Preparation}
	\label{sec: data preparation}
	
	Data is modelled and transformed on a profile level since the main objective is to group guests based on a set of attributes that define behaviour. The raw data, described in Section 3, is transformed into a data set containing 25 features. These features are constructed by making use of relationships between the tables, provided in Table 6 in Appendix B. Table 6 provides an overview of features, including the data type, possible values, and a description. In the table, six attributes are constructed based on business logic, described in the next section.
	
	\subsection{Business Logic}
	\label{subsec: business logic}
	
	A new set of features are attached to the data set based on existing features, a process known as feature engineering (Nargesian et al., 2017). These new features (1) guide the manual review process of the outcome of the algorithm; and (2) provide the model with additional context. The business logic, dictated by the hotel company, is constructed with the existing feature(s) that meet(s) a set of conditions on a profile level. These constructed features, which consider all data points attached to a profile, are binary values that indicate whenever the conditions are met. Table 3 provides an overview of the implemented business logic.
	
	\begin{table}[]
		\centering
		\begin{tabular}{l|lll}
			\multirow{2}{*}{\textbf{Name}} & \multicolumn{3}{c}{\textbf{Rule}}                      \\ \cline{2-4} 
			& \textbf{Feature} & \textbf{Condition} & \textbf{Value} \\ \hline
			Room Night Stay                & Length of Stay   & ==                 & 1              \\
			\multirow{2}{*}{Short Stay}    & Length of Stay   & \textgreater{}     & 1              \\
			& Length of Stay   & \textless{}=       & 3              \\
			Long Stay                      & Length of Stay   & \textgreater{}     & 3              \\
			Last-Minute Booker             & Lead Time        & \textless{}        & 3              \\
			Early-Bird Booker              & Lead Time        & \textgreater{}     & 45             \\
			Repeat Frequency Medium        & Repeat Count     & \textgreater{}     & 1              \\
			Loyalty Member                 & Loyalty          & is not             & NULL           \\ \hline
		\end{tabular}
	\caption{Overview of business logic}
	\end{table}
	
	\subsection{Dimensionality Reduction}
	\label{subsec: Dimensionality Reduction}
	
	The so-called curse of dimensionality can occur when a large number of features contain more than 15 distinct values (Hinneburg \& Keim, 1999). Dimensionality reduction is a method for encountering this phenomenon by transforming high-dimensional to low- dimensional space. Reducing the space of features also benefits the computational time of an algorithm.
	
	There are three general rules which are applied to reduce the space of features, dependent on the nature of the feature. The first rule impacts attributes that are a percentage of a total based on another attribute, e.g., the share of reservations made via a direct source. Even though the range of values is controlled, between 0 and 1, the number of unique values can still be high. This type is transformed by taking steps of $0.2$. The impact of this rule is that only six distinct values remain instead of 48 in the given example. The second rule is implemented to counter outliers is replacing values that exceed the $95\%$ quantile with the $95\%$ quantile. This rule is applied to features that are integers. This rule is not applied to revenue-related attributes, for which a third rule is introduced. Revenue-related features are rounded to the nearest $100$ and a maximum of $99\%$ quantile is set. Revenue is seen as one of the main characteristics for grouping guests and determining their (loyalty) status. Therefore, a different quantile range and rounding is used for these types of features. Features that are known to be binary values are not altered.
	
	\section{Model}
	\label{sec: model}
	
	For the cluster analysis, a machine learning technique is applied that divides guest profiles into different clusters. This algorithm, hierarchical clustering, detects patterns within an unlabelled data set, making use of minimal human input (Alpaydin, 2009). The unsupervised machine learning model is briefly explained first. Second, the evaluation criterion, which provides the optimal number of clusters, is described.
	
	\subsection{Hierarchical Clustering}
	\label{subsec: Hierarchical Clustering}
	
	Hierarchical clustering is an unsupervised machine learning algorithm that clusters data objects within a data set. An agglomerative (bottom-up) approach clusters data objects, with each data object constituting a cluster by itself as a start. This approach does not require a priori information about the number of clusters, which is instead determined by the evaluation criteria. An iterative process (1) identifies the most similar data objects; and (2) merges these data objects into a cluster. This process continues until a single cluster has been created.
	
	The identification of similar data objects in order to merge clusters requires a distance metric. This metric is used to compute all the pair-wise dissimilarities between observations in the data set. The data in Section 5 contains mixed data, a combination of numerical and categorical variables. Gower?s distance metric (Gower, 1971) can handle mixed data and is set as the distance metric.
	
	A linkage function merges clusters in an agglomerative approach based on the distance metric. Ward's minimum variance method (Ward, 1963) analyses cluster variances and minimizes the total within-cluster variance. This linkage function tends to generate balanced cluster sizes (Hands and Everitt, 1987) and is therefore robust against outliers (El-Hamdouchi and Willett, 1986). One of the main goals is to form well-balanced market segments, which is why Ward's method is chosen as the linkage function.
	
	\subsection{Evaluation Criterion}
	\label{subsec: Evaluation Criterion}
	
	Defining the number of clusters is a closely related challenge to clustering itself when clustering unlabelled data objects. With the application of hierarchical clustering, evaluation methods determine how well a model fits data objects. The elbow method (Thorndike, 1953) is applied to determine the optimal cluster number, given a range of cluster numbers.
	
	The elbow method calculates the sum of squares of a cluster number, defined as $k=1,...,20$. The point where the `elbow' appears in the slope of the sum of squares, from steep to shallow, determines the optimal cluster number. A disadvantage of this method is that two researches can point out a different `elbow' point. Therefore, a criterion is added, which is the sum of squares of $k$ divided by the sum of squares of $k=1$. This criterion, defined as $c$, is $100\%$ when the cluster number is $1$ and $0\%$ when all points are their own cluster. The first order difference of this criterion, defined as $d^{(1)} = c_{k} - c_{k-1}$, is calculated for each $k>1$. The second order difference of the criterion, defined as $d^{(2)}_{k} = d^{(1)}_{k} - d^{(1)}_{k-1}$, is calculated for each $k>2$. With this information, the \textit{relative strength} for cluster $k$ is calculated by $d^{(2)}_{k+1} - d^{(1)}_{k+1}/(k + 1)$ when $d^{(2)}_{k+1} > d^{(1)}_{k+1}$. The optimal cluster number is defined as the maximum relative strength. The relative strength is damped by dividing it by the number clusters since generally speaking more clusters are less meaningful.
	
	A computational limitation arises when classifying a large set of profiles. To overcome this limitation, multiple trails are executed with a subset of profiles to determine the optimal number of clusters. A trial with this optimal number of clusters is taken as a base, resulting that a subset of profiles have a cluster label attached. A k-nearest neighbours algorithm, with $k = 1$, is applied to classify the complete dataset.
	
	In addition, a conclusion is drawn about the stability of the outcome. If the optimum number of clusters is scattered over a large set of distinctive numbers, the result is seen as unstable since there is no distinctive optimum. A result is seen as extremely stable if the optimum is a single value for all trials. If the outcome is unstable, another set of attributes needs to be selected. For each of the 15 trials, a random selection of $10,000$ profiles serves as a sample. The number with the highest frequency is the optimal cluster number.
	
	\section{Results}
	\label{sec: results}
	
	This section describes the results of the hierarchical clustering model. First, the optimal number of clusters is determined. This number is used as input for the machine learning algorithm. The segments are labelled by the marketing department of the hotel after evaluating the characteristics of the clusters. Finally, segmentation over time is analysed and insights are presented about profiles transitioning from one segment to another.
	
	Defining the number of clusters is a challenge in the application of an unsupervised machine learning algorithm. Section 6.2 has described the method for determining the optimal number of clusters. Out of 15 executions, eight times 8 clusters were identified as the optimal number. Figure 3 shows a trial where 8 is identified as optimum. The left y-axis represents the criterion and the right y-axis represents the relative strength over a range of clusters. The relative strength is the highest eight clusters, despite the identification of six `elbow' points. Appendix C contains Table 7, which shows the numerical result of the trial shown in Figure 3. The average number of seconds to complete one executions was $3.4$ seconds with a standard deviation of $0.2$ seconds.
	
	\begin{figure}[]
		\centering
		\includegraphics[width=.6\linewidth]{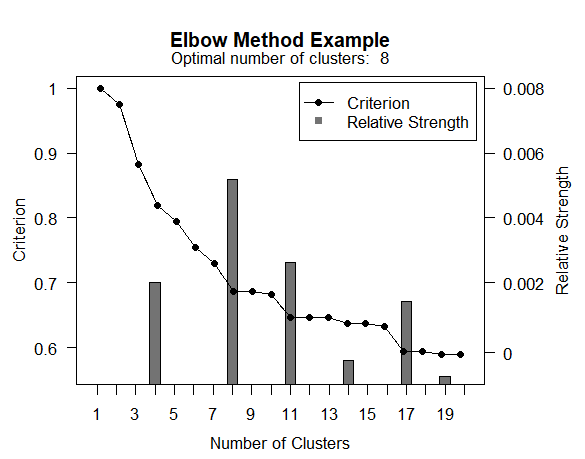}
		\caption{Example of the elbow method}
		\label{fig:Results - Elbow Method example}
	\end{figure}
	
	Given that 8 is the optimal cluster number, the hierarchical clustering algorithm is executed on $170,000$ profiles with data from 01-01-2016 until 01-01-2020. Each cluster is labelled with a name defined by the hotel company, based on an attribute evaluation.  A detailed overview of attributes is displayed in Table 8 in Appendix D, where characteristics per cluster are highlighted. The segments are defined as follows:
	
	\begin{itemize}
		\item \textbf{GIT} ($28.01\%$): A Group Inclusive Tour (GIT) consisting of individuals who buy a group package and travel with others on a preset itinerary. They are mainly cruise guests, who book through an agent instead of directly through the hotel. Typically, these guests stay once, so no marketing strategy will be set up.
		\item \textbf{GDS} ($13.70\%$): Mainly corporate and transient bookings via agencies, which is the typical profile of bookings via the Global Distribution System (GDS). These guests do not book for themselves but through an agent, either for their corporate stay with a contracted rate, or through consortia. As they book  trough an agent, it is difficult to reach them directly. For example, companies have their own fixed way of booking. Activating this group with book direct campaigns does not work.
		\item \textbf{Cancelled - Direct} ($3.89\%$): Guests who booked directly with the hotel, but cancelled. In the cancellation email, guests are stimulated to book their next stay by booking directly with the hotel.
		\item \textbf{Transient - Direct} ($11.92\%$): These are leisure bookings created through \textit{direct} channels and consists of stays (mostly) during the weekends. At the moment, this is the main focus of the marketing department.
		\item \textbf{MICE} ($16.51\%$): Guests who stay at the hotel during Meetings, Incentives, Conferences or Exhibitions (MICE). If guests book directly via a promo code, there is a upsell possibility in the booking engine. In addition, active promotions of the hotel for leisure purposes to the participants of events will be set up.
		\item \textbf{OTA} ($6.26\%$): Guests who book via in-direct channels. It is important to obtain their direct email address instead of an email address generated by third-party channels. By obtaining their email address, there is the possibility to target those guests to book with the hotel directly for their next stay and restau rants/spa facilities will be promoted.
		\item \textbf{Cancelled - Indirect} ($9.66\%$): Guests who book via \textit{indirect} channels, but who have cancelled. Depending on the channel they used, a cancellation email is sent. In the cancellation email, guests are stimulated to book their next stay by booking directly with the hotel.
		\item \textbf{Repeat} ($10.07\%$): Guest who are identified as recurring. There is a high percentage of repeat guests, but around $50\%$ book directly. For marketing, it would be beneficial to organize a campaign to make sure those guests will use the hotel channel next time instead of a third-party channel. This can be done during check-out, an email campaign, pre-arrival mail or post-stay mail.
	\end{itemize}
	
	\subsection{Over Time}
	\label{subsec: over time}
	
	Profiles are evaluated on several timestamps, 01-01-2016, 01-01-2017, 01-01-2018, 01- 01-2019 and 01-01-2020. The machine learning model trained on timestamp 01-01-2020 is used to evaluate profiles for the other timestamps. The temporal analysis provides an understanding of the changes in profile attributes that cause a transition from one segment to another. The result guides marketeers on what actions need to be taken to make profiles transition from one segment to another. This transition can either be positive ? profiles changing to high profitable segments ? or negative ? profiles changing to less profitable segments, where profitable is defined as average amount spend per stay. By understanding the actions that need to be taken or avoided, the marketing budget can be optimized. For each of these timestamps, the attributes are constructed using data up to that moment in time.
	
	Figure 4 is a flow diagram that visualizes transitions for each timestamp, where the width of each flow represents its quantity. The trained model divides profiles into 8 segments; however, an additional stream is added to each timestamp to allow for new guests. These new guests have a moment of arrival for their reservation between two timestamps. The transition flow is only visualized if it exceeds a threshold of $0.1\%$ profiles at that timestamp.
	
	\begin{figure}[H]
		\centering
		\includegraphics[width=1\linewidth]{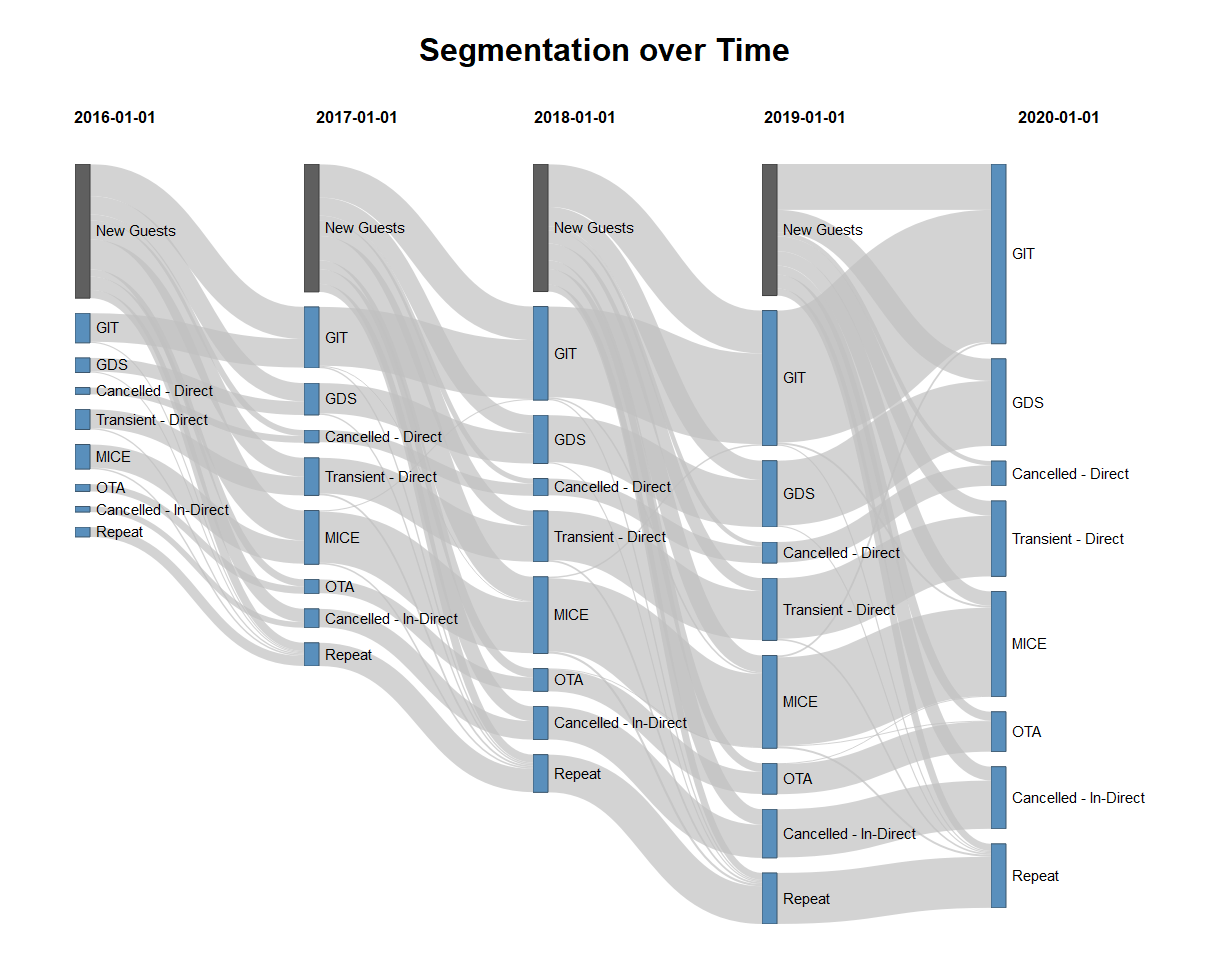}
		\caption{Segmentation over Time}
		\label{fig:Results - Segmentation over Time}
	\end{figure}
	
	Table 4 presents the number of profiles for each segment from Figure 4. The \textit{Total Profiles} row does not include \textit{New Guests} at that timestamp since these are evaluated the next timestamp. The segment \textit{GIT}, \textit{GDS}, \textit{Cancelled - Indirect}, and \textit{Repeat} grow over time in terms of percentage. The largest growth is for \textit{GIT} both in terms of percentage points and the absolute number of profiles.
	
	\begin{table}[H]
		\centering
		\addtolength{\leftskip} {-2cm}
		\addtolength{\rightskip}{-2cm}
		\begin{tabular}{l|ccccc}
			\textbf{Segment}        & \textbf{01-01-2016} & \textbf{01-01-2017} & \textbf{01-01-2018} & \textbf{01-01-2019} & \textbf{01-01-2020} \\ \hline
			New Guests              & 35544               & 33880               & 33789               & 34852               & -                   \\ \hline
			GIT                     & 8111                & 16360               & 24964               & 35996               & 47899               \\
			GDS                     & 3997                & 8575                & 13021               & 17823               & 23425               \\
			Cancelled - Direct      & 1948                & 3467                & 4624                & 5649                & 6648                \\
			Transient - Direct      & 5495                & 10155               & 13726               & 16820               & 20379               \\
			MICE                    & 6813                & 14517               & 20639               & 24716               & 28229               \\
			OTA                     & 2235                & 3980                & 6262                & 8397                & 10711               \\
			Cancelled - Indirect    & 1655                & 5056                & 8933                & 12949               & 16512               \\
			Repeat                  & 2697                & 6385                & 10206               & 13814               & 17213               \\ \hline
			\textbf{Total Profiles} & \textbf{32951}      & \textbf{68495}      & \textbf{102375}     & \textbf{136164}     & \textbf{171016}    
		\end{tabular}
		\caption{Number of profiles per segment for each time stamp}
	\end{table}
	
	Table 5 presents the \textit{New Guest} profiles from Figure 4 and their assignment to a segment. The number of new profiles over the year remains consistent, with a standard deviation of $2.4\%$ based on the average number of new guests per year. The main reason why \textit{GIT} sees the largest growth is due to the increasing number of new guests assigned to it rather than transitioning of profiles. A shift takes place for new guest assigned to \textit{MICE} : this decreases over time, from $22.43\%$ in 01-01-2016 to $11.04\%$ in 01-01-2020.
	
	\begin{table}[H]
	\centering
	\addtolength{\leftskip} {-2cm}
	\addtolength{\rightskip}{-2cm}
		\begin{tabular}{l|ccccc}
			\textbf{Segment}            & \textbf{01-01-2016} & \textbf{01-01-2017} & \textbf{01-01-2018} & \textbf{01-01-2019} & \textbf{01-01-2020} \\ \hline
			GIT                         & -                   & 8517                & 8893                & 11366               & 12160               \\
			GDS                         & -                   & 4861                & 4787                & 5123                & 5884                \\
			Cancelled - Direct          & -                   & 1584                & 1347                & 1152                & 1132                \\
			Transient - Direct          & -                   & 4994                & 4025                & 3425                & 3935                \\
			MICE                        & -                   & 7973                & 6446                & 4477                & 3818                \\
			OTA                         & -                   & 1890                & 2335                & 2282                & 2381                \\
			Cancelled - Indirect        & -                   & 3525                & 3916                & 4134                & 3665                \\
			Repeat                      & -                   & 2200                & 2131                & 1830                & 1877                \\ \hline
			\textbf{Total New Profiles} & \textbf{-}          & \textbf{35544}      & \textbf{33880}      & \textbf{33789}      & \textbf{34852}      \\ \hline
		\end{tabular}
		\caption{Assignment of \textit{New Guests} segment for each time stamp}
	\end{table}
	
	Two segments, \textit{Transient - Direct} and \textit{Repeat}, are of special interest to the hotel. For each of these, a flow diagram is created (Figure 5 and Figure 6) which highlights, per timestamp, the origin of the profile that transitions to the segment of interest. The composition of both segments is reliant on the group of profiles that transition in each year.
	
	\newpage
	
	\begin{figure}[H]
		\centering
		\includegraphics[width=.97\linewidth]{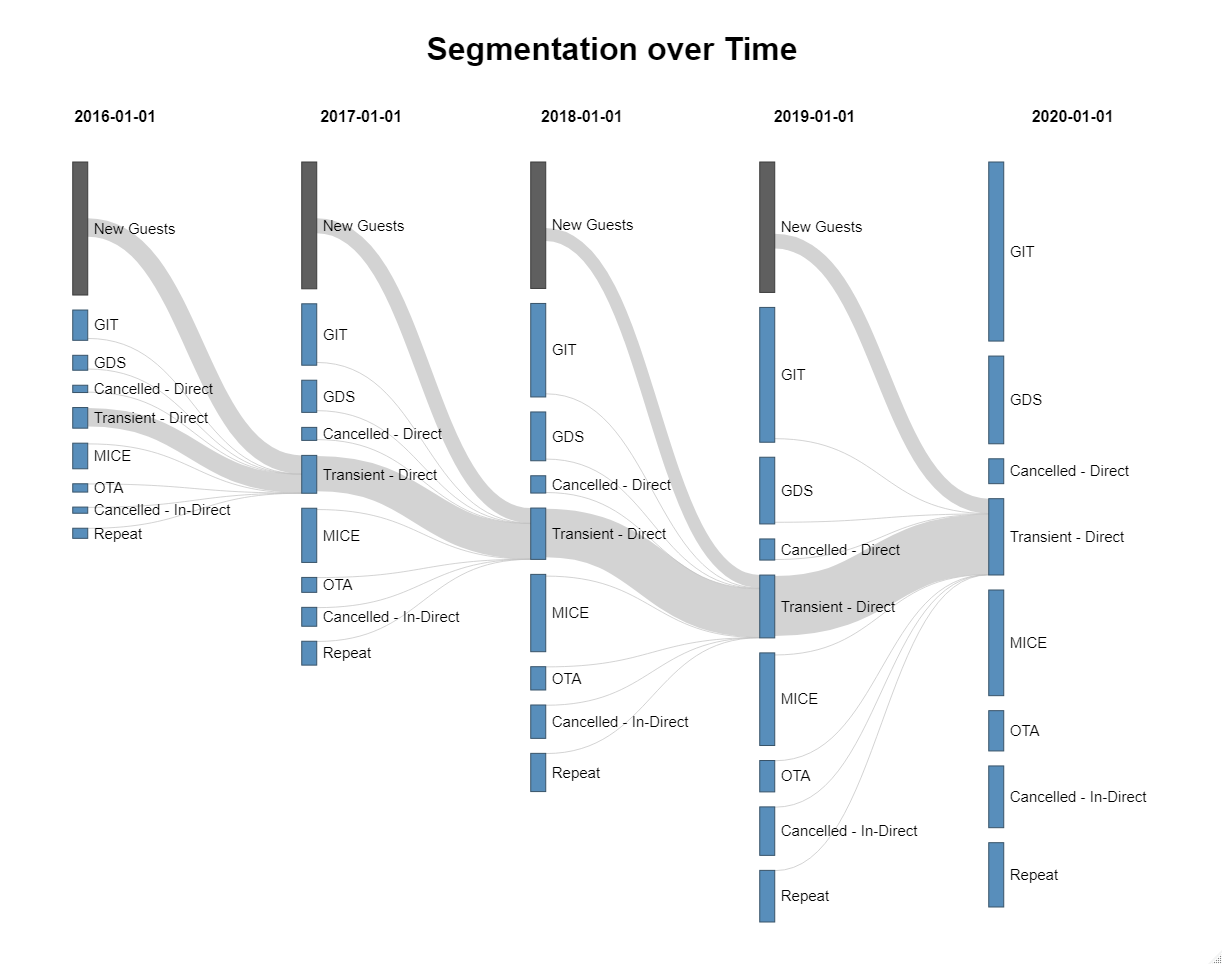}
		\caption{ Segmentation over Time for \textit{Transient - Direct}}
		\label{fig:Results - Segmentation over Time Cluster 4}
	\end{figure}

	\begin{figure}[H]
		\centering
		\includegraphics[width=.97\linewidth]{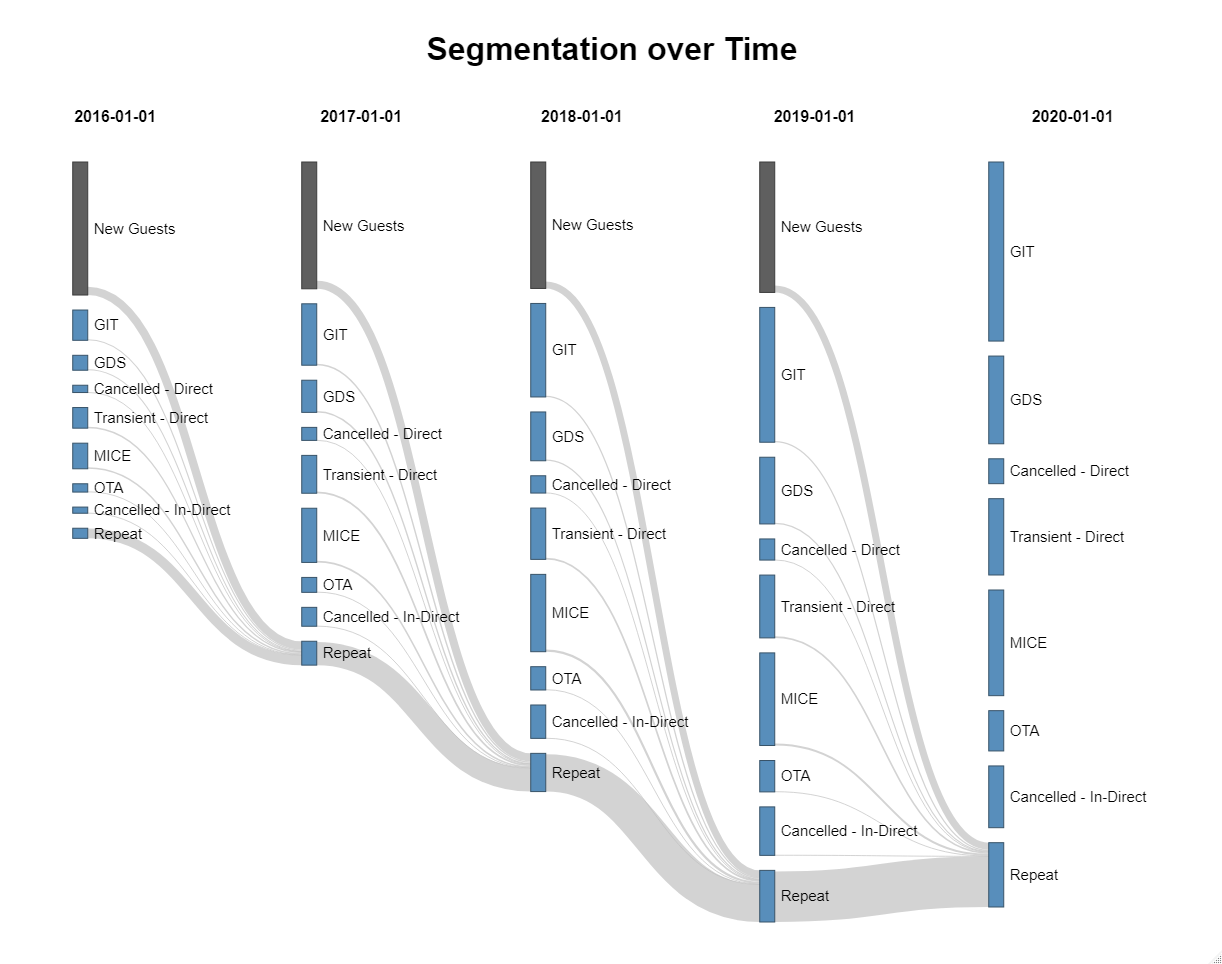}
		\caption{Segmentation over Time for \textit{Repeat}}
		\label{fig:Results - Segmentation over Time Cluster 8}
	\end{figure}

	\newpage
	
	New guests are the main source of profiles that transition to the segments of interest every year. For \textit{Transient - Direct}, the most popular transition is from \textit{GDS} (an average of $1.15\%$ of the segment) and \textit{MICE} (an average of $0.35\%$ of the segment). Regarding \textit{Repeat}, the most popular transitions are from \textit{Transient - Direct} ($4.25\%$ average) and \textit{MICE} ($3.34\%$ average). Guest are more likely to transition to \textit{Repeat} than to \textit{Transient - Direct}. The main reason for this transition is the positive change in features \textit{ReservationTotal} (average $\Delta+1.04$), \textit{RevenueTotal} (average $\Delta+115.2\%$), and \textit{RepeatTotal} (average $\Delta+0.94$). This specific change in features explains the purpose of the transition. The change in \textit{RepeatTotal} is not larger than 1 which is because there is a small percentage of guests that is already a return guest. The combination of features, rather than a change in a single attribute, causes the transition.
	
	\subsection{Activation}
	\label{subsec: Activation}
	The guests with a reservation at this hotel can be roughly divided into business travellers and leisure travellers by the hotel company. The sales department is responsible for the business-to-business (B2B) side, while the marketing department focuses on the business-to-customer (B2C) side. Due to this division of roles between the departments, a specific selection of the hierarchical clustering results will be activated with a grouped marketing strategy instead of a pooled one. The \textit{Transient - Direct} and \textit{Repeat} segments are main focus of the marketing department.
	
	The strategy of the marketing department will change from a generic newsletter to newsletters that are targeted to guests based on their stays, interests, and preferences. This type of newsletter is sent via email and is seen as direct marketing to the guest, where direct marketing consents are given. Due to the General Data Protection Regulation (GDPR), a hotel is not able to opt-in all guests for direct marketing consent. Within the data set, $47.6\%$ are opt-in guests for direct email marketing. Even though sending out an email does not cost anything extra, time restricts resources for creating an unlimited number of different newsletters.
	
	In addition to direct email marketing, the hotel company will use the segmentation result for online campaigns on different platforms such as Google Ads, Instagram, and Facebook. These campaigns will be set up to attract new guests after analysing reservation and profile information such as the country of origin, booking lead time, and packages chosen of segments from interest. Online marketing campaigns become data-driven by targeting similar guests belonging to valuable segments at the right time with a certain promotion strategy.
	
	\section{Conclusion \& Discussion}
	\label{sec: conclusion}
	
	This study provides a guideline on how hospitality can make use of unsupervised machine learning methods to segment guests and develop a marketing strategy. The attributes of a profile, which serve as the input for the algorithm, need to reflect the underlying business question, which needs to be defined by the hotel company itself. Otherwise, there is the risk of a low adoption rate of the result by the end user. The outcome can be evaluated by the hotel company and will impact its marketing strategy, which shifts towards a more personalized approach.
	
	Based on the available data, features are constructed on a profile level, that reflect which aspects the hotel wants to use to segment its guests. The data analysis shows that attributes can divide guests based on spending; for example, the average amount spent by repeat guests is $16.7\%$ higher than other guests. Binary values, which link reservation information to a group, company, and agency, show probabilities of some- one becoming a returning guest. Demographic information, such as age and country of origin, is excluded from the segmentation process on purpose and used in the post-hoc phase of creating personalised marketing strategies.
	
	Hierarchical clustering (an agglomerative approach) in combination with the elbow method results in a split of profiles that the hotel company classified. The model was configured with Gower's distance metric to determine similarities between profiles and used the Ward's minimum variance method to merge clusters. The optimal number of clusters was eight, each cluster distinctive enough for the hotel company to provide a label and description.
	
	Each segment needs to meet six criteria to evaluate the outcome of any segmentation method, as described in Section 2. Each segment has at least a combination of two to four characteristics; therefore, the result meets the identifiability criterion. The segment sizes vary between $3.89\%$ and $28.01\%$ of the total number of profiles. Even though the smallest segment is $3.89\%$, it is distinguished by a set of characteristics and recognized by the hotel as a separate group of guests. Therefore, the substantiality criterion is met. As stated in Section 7.2, guests are reached via direct email marketing if consent is given, which meets the accessibility criterium. Section 7.1 provides evidence that each segment is stable over five years. The marketing department provided an initial global strategy per segment, which confirms the actionability criterion. The responsiveness criterion could not be checked since no data was available about the number of received emails, opens, and clicks from direct email campaigns.
	
	The transitions of profiles over time provide insight into how guests naturally transition from one segment to another. This analysis shows that, on average, $4.35\%$ of the profiles made the transition from \textit{Transient - Direct} to \textit{Repeat}. Marketing departments gain understanding through the reasons why guests made this transition. Analysing the profiles that made this transition provides knowledge that helps them to decide how to encourage the transition to a more profitable segment by analysing the profiles that made this transition.
	
	There is a continuous inflow of new guests; however, there is no particular outflow of guests. One can wonder whether marketing is still efficient when a profile is classified in a segment of interest, but the last stay or interaction was many years ago. Figures 5 and 6 show that the size of a segment on a time is reliant on the customer base of the previous timestamp. For example, guests flow to a new segment on a timestamp, for example \textit{Guest - Outflow}, when meeting a criteria (e.g. last email interaction was 5 years ago, last made reservations was 3 years ago). By updating the customer base, the hotel will target an active customer base.
	
	Point-of-sale data points can be brought in to complete the lifetime value of a guest. In this case, monetary information is one of the main drivers to segment guests. The hotel company has a unique revenue composition, close to $50\%$ of the total revenue is seen as ancillary. There was no possibility to connect these data points to profiles. By bringing in these point-of-sale data points, profiles can be evaluated on their total lifetime value.
	
	The hierarchical clustering model that provided the result based on all data points is used to classify profiles for multiple timestamps, see Section 7.1. The possibility  of change in cluster composition is not taken into account since the hotel chain was interested in their current composition. Guests change over time, which means there is a realistic chance that clusters split and merge. The consequence is that marketing departments should evaluate a new trained hierarchical clustering model on a recurring basis to identify if new clusters appear and/or clusters merged.
	
	\newpage
	
	\nocite{1, 2, 3, 4, 5, 6, 7, 8, 9, 10, 11, 12, 13, 14, 15, 16, 17, 18, 19, 20, 21}
	\bibliographystyle{plain}
	\bibliography{bibtex} 
	
	\newpage
	
	\appendix
	\newpage
	\section{Data Structure} 
	\label{sec: Appendix - Data Structure}
	
	An overview of the available data, the linkage structure, and data fields is provided in Figure 7. The data originates from the Property Management System (PMS), where individuals can have multiple reservations and where reservations can have multiple folios. Each table contains three columns: key, name, and type. The key column indicates whether there is a primary key (PK) or a foreign key (FK). The foreign key is the column in a table that refers to the primary key in another table.
	
	\begin{figure}[H]
		\centering
		\includegraphics[width=1\linewidth]{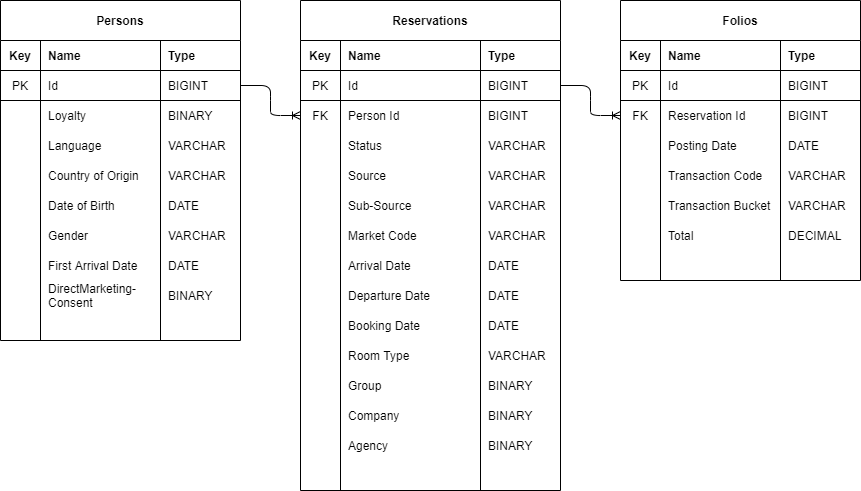}
		\caption{Database structure including relationships and fields per database table}
		\label{fig:Data - Database Structure}
	\end{figure}
	
	\newpage
	
	\section{Attributes Overview} 
	\label{sec: Appendix - Attributes Overview}
	
	An overview of the attributes, and the corresponding details, used for the hierarchical clustering algorithm to group profiles

	\begin{table}[H]
		\centering
		\addtolength{\leftskip} {-2cm}
		\addtolength{\rightskip}{-2cm}
		\begin{tabular}{l|lllp{8cm}|}
			\textbf{Id} & \textbf{Name}               & \textbf{Type} & \textbf{Values}  & \textbf{Description}                                                                                                                                         \\ \hline
			1           & ReservationsTotal           & Integer       & \textgreater 0   & Sum of reservation a profile created                                                                                                                         \\
			2           & ReservationsHistoric        & Decimal       & {[}0, 1{]}       & Percentage of \textit{ReservationsTotal} with status \textit{Historic}                                                                                                         \\
			3           & ReservationsCancelled       & Decimal       & {[}0,1{]}        & Percentage of \textit{ReservationsTotal} with status \textit{Cancelled}                                                                                                        \\
			4           & ReservationsCompany         & Decimal       & {[}0, 1{]}       & Percentage of \textit{ReservationsTotal} associated to a company                                                                                                     \\
			5           & ReservationsAgency          & Decimal       & {[}0, 1{]}       & Percentage of \textit{ReservationsTotal} associated to a agency                                                                                                       \\
			6           & ReservationsGroup           & Decimal       & {[}0, 1{]}       & Percentage of \textit{ReservationsTotal} associated to a group                                                                                                        \\
			7           & ReservationsSourceDirect    & Decimal       & {[}0, 1{]}       & Percentage of \textit{ReservationsTotal} made via a \textit{Direct} source                                                                                                     \\
			8           & ReservationsSourceIndirect & Decimal       & {[}0, 1{]}       & Percentage of \textit{ReservationsTotal} made via an \textit{Indirect} source                                                                                                \\
			9           & RevenueTotal                & Integer       & \textgreater{}=0 & Sum of total of folios                                                                                                                                       \\
			10          & RevenueAverage              & Integer       & \textgreater{}=0 & \textit{RevenueTotal} divided by \textit{ReservationsTotal}                                                                                                                   \\
			11          & RevenueTotalRoom            & Integer       & \textgreater{}=0 & Sum of total of folios associated to TransactionClassification \textit{Room Revenue}                                                                                  \\
			12          & RevenueTotalAncillary       & Integer       & \textgreater{}=0 & Sum of total of folios associated to TransactionClassification \textit{Ancillary Revenue}                                                                             \\
			13          & RepeatBinary                & Binary        & \{0, 1\}         & \textit{ReservationsTotal} \textgreater 1                                                                                                                            \\
			14          & Repeat Total                & Integer       & \textgreater{}=0 & \textit{ReservationsTotal} - 1                                                                                                                                       \\
			15          & RepeatFrequencyMediumBinary & Binary        & \{0,1\}          & \textit{ReservationsTotal} and \textit{RepeatBinary} = 1                                                                                                                      \\
			16          & RepeatLast365Binary         & Binary        & \{0,1\}          & Has arrival date of reservation within last 365 days from evaluation date                                                                                    \\
			17          & WeekStay                    & Decimal       & {[}0, 1{]}       & Percentage of \textit{ReservationsTotal} where length of stay \textless{}5 and day of week of arrival date and day of week of departure date Mon, Tue, Wed, Thu, Fri \\
			18          & WeekendStay                 & Decimal       & {[}0, 1{]}       & Percentage of \textit{ReservationsTotal} where length of stay \textless{}3 and day of week of arrival date and day of week of departure date is Fri, Sat, Sun        \\
			19          & LOSAverage                  & Decimal       & \textgreater{}=0 & Average of length of stay (number of days between arrival date and departure date) of all reservations                                                       \\
			20          & SingleNightBinary           & Binary        & \{0, 1\}         & \textit{LOSAverage} = (1, 3)                                                                                                                                          \\
			21          & ShortStayBinary             & Binary        & \{0, 1\}         & \textit{LOSAverage} \textgreater{}3                                                                                                                                   \\
			22          & MediumStayBinary            & Binary        & \{0, 1\}         & \textit{LOSAverage} = 1                                                                                                                                               \\
			23          & LastMinuteBookerBinary      & Binary        & \{0, 1\}         & \textit{LeadTimeAverage} (average number of days between booking date and arrival date of reservations) \textless{}=3                                                 \\
			24          & EarlyBirdBookerBinary       & Binary        & \{0,1\}          & \textit{LeadTimeAverage} (average number of days between booking date and arrival date of reservations) \textgreater{}= 45                                            \\
			25          & Loyalty Binary              & Binary        & \{0, 1\}         & Profile loyalty level is not NULL                                                                                                                            \\ \hline
		\end{tabular}
		\caption{Profile attributes overview}
	\end{table}
	
	\newpage
	
	\section{Elbow Method Example} 
	\label{sec: Appendix - Elbow Method Example}
	
	This section showcases one of the trails that was executed to determine the optimal number of clusters for the hierarchical clustering algorithm. Table 7 is a numeric representation of Figure 3. The formulas to calculate these values are described in Section 6.2.
	
	\begin{table}[H]
		\centering
		\addtolength{\leftskip} {-2cm}
		\addtolength{\rightskip}{-2cm}
		\begin{tabular}{l|lp{2cm}p{2.5cm}lp{2cm}}
			\textbf{Cluster} & \textbf{Criterion} & \textbf{First Order Difference} & \textbf{Second Order Difference} & \textbf{Elbow Binary} & \textbf{Relative Strength} \\ \hline
			1                & 1                  & -                               & -                                & -                     & -                          \\
			2                & 0.975              & 0.025                           & -                                & -                     & -                          \\
			3                & 0.883              & 0.092                           & -0.067                           & 0                     & 0                          \\
			4                & 0.82               & 0.063                           & 0.029                            & 1                     & 0.003                      \\
			5                & 0.794              & 0.026                           & 0.037                            & 0                     & 0                          \\
			6                & 0.755              & 0.039                           & -0.013                           & 0                     & 0                          \\
			7                & 0.73               & 0.025                           & 0.014                            & 0                     & 0                          \\
			8                & 0.686              & 0.044                           & -0.019                           & 1                     & 0.006                      \\
			9                & 0.686              & 0                               & 0.044                            & 0                     & 0                          \\
			10               & 0.682              & 0.004                           & -0.004                           & 0                     & 0                          \\
			11               & 0.646              & 0.036                           & -0.032                           & 1                     & 0.003                      \\
			12               & 0.646              & 0                               & 0.036                            & 0                     & 0                          \\
			13               & 0.646              & 0                               & 0                                & 0                     & 0                          \\
			14               & 0.637              & 0.009                           & -0.009                           & 1                     & 0.001                      \\
			15               & 0.637              & 0                               & 0.009                            & 0                     & 0                          \\
			16               & 0.632              & 0.005                           & -0.005                           & 0                     & 0                          \\
			17               & 0.594              & 0.038                           & -0.033                           & 1                     & 0.002                      \\
			18               & 0.594              & 0                               & 0.038                            & 0                     & 0                          \\
			19               & 0.59               & 0.004                           & -0.004                           & 1                     & 0                          \\
			20               & 0.59               & 0                               & 0.004                            & 0                     & 0                         
		\end{tabular}
		\caption{Trial result of applying the elbow method}
	\end{table}
	
	\newpage
	
	\section{Results of Hierarchical Clustering by Attributes} 
	\label{sec: Appendix - Results of Hierarchical Clustering by Attributes}
	
	Table 8 provides a detailed overview of the characteristics, by a grey cell color, of the segments presented in Section 7 based on 170, 000 profiles evaluated on 01-01-2020. Given the attributes overview in Appendix B, for decimal or integer type values, the average is calculated. Except for \textit{revenue} related attributes, these are transformed to a percentage of the average instead of absolute values and the \textit{RevenueAverage} is taken out. For binary-type attributes, a split is made between the \textit{True/False} values and the frequency percentage is given
	
	\begin{table}[H]
		\centering
		\addtolength{\leftskip} {-4cm}
		\addtolength{\rightskip}{-1cm}
		\begin{tabular}{ll|llp{1.2cm}p{1.2cm}llp{1.2cm}ll}
			&                    & \multicolumn{9}{c}{\textbf{Cluster}}                                                                                                      \\ \cline{3-11} 
			\textbf{Attribute}                      & \textbf{Catergory} & GIT   & GDS   & Cancelled - Direct & Transient - Direct & MICE   & OTA    & Cancelled - In-Direct & \multicolumn{1}{l|}{Repeat} & Overall \\ \hline
			ReservationsTotal                       &                    & 1.07  & 1.06  & 1.07               & 1.08               & 1.09   & 1.18   & 1.14                  & \multicolumn{1}{l|}{3.56}   & 1.34    \\
			ReservationHistoric                     &                    & 98.41 & 98.39 & 1.39               & 98.77              & 98.16  & 97.56  & 0.3                   & \multicolumn{1}{l|}{87.12}  & 83.98   \\
			ReservationCancelled                    &                    & 1.45  & 1.4   & \cellcolor{grey}85.57              & 1.09               & 1.46   & 2.16   & \cellcolor{grey}90.62                 & \multicolumn{1}{l|}{11.88}  & 14.36   \\
			ReservationCompany                      &                    & 13.36 & 14.71 & 42.18              & 38.61              & \cellcolor{grey}42.89  & 24.71  & 10.47                 & \multicolumn{1}{l|}{30.58}  & 24.78   \\
			ReservationAgency                       &                    & 60.44 & 56.02 & 24.74              & 11.1               & 19.87  & 42.66  & \cellcolor{grey}97.86                 & \multicolumn{1}{l|}{32.86}  & 45.38   \\
			ReservationGroup                        &                    & 40.94 & 12.82 & 49.35              & 39.78              & \cellcolor{grey}68.73  & 42.91  & 7.98                  & \multicolumn{1}{l|}{24.35}  & 37      \\
			ReservationSourceDirect                 &                    & 19.24 & 7.04  & \cellcolor{grey}99.55              & \cellcolor{grey}89.2               & \cellcolor{grey}87     & 37.3   & 1                     & \multicolumn{1}{l|}{49.76}  & 42.73   \\
			ReservationSourceIndirect              &                    & \cellcolor{grey}80.76 & \cellcolor{grey}92.96 & 0.45               & 10.8               & 13     & \cellcolor{grey}62.7   & \cellcolor{grey}99                    & \multicolumn{1}{l|}{50.24}  & 57.27   \\
			RevenueTotal                            &                    & 86.51 & 46.61 & 0.37               & 56.27              & 120.01 & 211.43 & 0.02                  & \multicolumn{1}{l|}{\cellcolor{grey}279.26} & 100     \\
			RevenueTotalRoom                        &                    & 89.89 & 46.9  & 0.34               & 48.11              & 122.52 & 227.14 & 0.01                  & \multicolumn{1}{l|}{264.21} & 100     \\
			RevenueTotalAncillary                   &                    & 74.81 & 45.28 & 0.47               & 83.51              & 112.33 & 158.99 & 0.05                  & \multicolumn{1}{l|}{330.02} & 100     \\
			\multirow{2}{*}{RepeatBinary}           & FALSE              & 95.71 & 94.08 & 95.46              & 91.4               & 93.15  & 91.74  & 98.93                 & \multicolumn{1}{l|}{14.61}  & 86.25   \\
			& TRUE               & 4.29  & 5.92  & 4.54               & 8.6                & 6.85   & 8.26   & 1.07                  & \multicolumn{1}{l|}{\cellcolor{grey}85.39}  & 13.75   \\
			RepeatTotal                             &                    & 0.06  & 0.07  & 0.05               & 0.11               & 0.11   & 0.18   & 0.01                  & \multicolumn{1}{l|}{2.55}   & 0.33    \\
			\multirow{2}{*}{RepeatFrequencyMedium}  & FALSE              & 99.05 & 99.17 & 99.99              & 98.73              & 98.36  & 97.52  & 99.99                 & \multicolumn{1}{l|}{\cellcolor{grey}57.02}  & 94.63   \\
			& TRUE               & 0.95  & 0.83  & 0.01               & 1.27               & 1.64   & 2.48   & 0.01                  & \multicolumn{1}{l|}{42.98}  & 5.37    \\
			\multirow{2}{*}{RepeatLast365Binary}    & FALSE              & 97.92 & 98.05 & 99.69              & 97.49              & 97.19  & 95.28  & 99.99                 & \multicolumn{1}{l|}{28.91}  & 90.77   \\
			& TRUE               & 2.08  & 1.95  & 0.31               & 2.51               & 2.81   & 4.72   & 0.01                  & \multicolumn{1}{l|}{\cellcolor{grey}71.09}  & 9.23    \\
			WeekStay                                &                    & 39.24 & 49.67 & 55.24              & 53.93              & 47.74  & 30.79  & 41.45                 & \multicolumn{1}{l|}{51.82}  & 45.33   \\
			WeekendStay                             &                    & 34.29 & \cellcolor{grey}48.92 & 23.9               & 44.82              & 20.29  & 4.18   & 29.46                 & \multicolumn{1}{l|}{31.82}  & 32      \\
			LOSAverage                              &                    & 2.41  & 1.08  & 2.19               & 1.09               & 2.67   & 4.26   & 2.58                  & \multicolumn{1}{l|}{1.94}   & 2.21    \\
			\multirow{2}{*}{SingleRoomNightBinary}  & FALSE              & 97.14 & 6.56  & 60.47              & 7.99               & 96.81  & 98.51  & 64.92                 & \multicolumn{1}{l|}{69.79}  & 66.59   \\
			& TRUE               & 2.86  & 93.44 & 39.53              &\cellcolor{grey} 92.01              & 3.19   & 1.49   & 35.08                 & \multicolumn{1}{l|}{30.21}  & 33.41   \\
			\multirow{2}{*}{ShortStayBinary}        & FALSE              & 9.13  & 94.25 & 54.28              & 93.79              & 19.72  & 77.7   & 55.91                 & \multicolumn{1}{l|}{40.98}  & 47.37   \\
			& TRUE               & \cellcolor{grey}90.87 & 5.75  & 45.72              & 6.21               & \cellcolor{grey}80.28  & 22.3   & 44.09                 & \multicolumn{1}{l|}{59.02}  & 52.63   \\
			\multirow{2}{*}{MediumStayBinary}       & FALSE              & 93.75 & 99.51 & 85.45              & 99.24              & 83.5   & 23.79  & 79.22                 & \multicolumn{1}{l|}{89.83}  & 86.29   \\
			& TRUE               & 6.25  & 0.49  & 14.55              & 0.76               & 16.5   & \cellcolor{grey}76.21  & 20.78                 & \multicolumn{1}{l|}{10.17}  & 13.71   \\
			\multirow{2}{*}{LastMinuteBookerBinary} & FALSE              & 90.82 & 75.97 & 91.26              & 77.76              & 92.34  & 92.3   & 92.84                 & \multicolumn{1}{l|}{86.11}  & 87.27   \\
			& TRUE               & 9.18  & 24.03 & 8.74               & 22.24              & 7.66   & 7.7    & 7.16                  & \multicolumn{1}{l|}{13.89}  & 12.73   \\
			\multirow{2}{*}{EarlyBirdBinary}        & FALSE              & 81.64 & 87.43 & 72.22              & 86.2               & 81.51  & 76.75  & 66.94                 & \multicolumn{1}{l|}{78.99}  & 80.57   \\
			& TRUE               & 18.36 & 12.57 & 27.78              & 13.8               & 18.49  & 23.25  & 33.06                 & \multicolumn{1}{l|}{21.01}  & 19.43   \\
			\multirow{2}{*}{LoyaltyLevelBinary}     & FALSE              & 99.32 & 99.09 & 99.84              & 99.07              & 99.39  & 98.85  & 99.88                 & \multicolumn{1}{l|}{94.69}  & 98.83   \\
			& TRUE               & 0.68  & 0.91  & 0.16               & 0.93               & 0.61   & 1.15   & 0.12                  & \multicolumn{1}{l|}{5.31}   & 1.17   
		\end{tabular}
		\caption{Attribute overview per segment}
	\end{table}

\end{document}